\documentclass{article} % For LaTeX2e
\usepackage{iclr2026_conference,times}

\usepackage{amsmath,amsfonts,bm}

\def\eqref#1{equation~\ref{#1}}
\def\1{\bm{1}}

\DeclareMathAlphabet{\mathsfit}{\encodingdefault}{\sfdefault}{m}{sl}
\SetMathAlphabet{\mathsfit}{bold}{\encodingdefault}{\sfdefault}{bx}{n}

\usepackage{hyperref}
\usepackage{url}
\usepackage{times}
\usepackage{amsmath,amssymb,amsfonts}
\usepackage{graphicx}
\usepackage{booktabs}
\usepackage{algorithm}
\usepackage{algorithmic}
\usepackage{hyperref}
\usepackage{xcolor}
\usepackage{colortbl}
\usepackage{multirow}
\usepackage{subcaption}
\usepackage{tikz}
\usepackage{float}
\usetikzlibrary{shapes.geometric, arrows.meta, positioning, calc, fit, backgrounds, shadows}

\title{Adaptive Loops and Memory in Transformers: Think Harder or Know More?}

\author{Markus Frey\textsuperscript{1, 2, 3}, 
Behzad Shomali\textsuperscript{1, 3}, 
Ali Hamza Bashir\textsuperscript{1, 2}, 
David Berghaus\textsuperscript{1, 2},\\
\bf Joachim Koehler\textsuperscript{1, 2},
Mehdi Ali\textsuperscript{1, 2}
\\
Lamarr Institute\textsuperscript{1}, Fraunhofer IAIS\textsuperscript{2}, University of Bonn\textsuperscript{3}
\\
\small \texttt{markus.frey@iais.fraunhofer.de} 
}

\iclrfinalcopy % Uncomment for camera-ready version, but NOT for submission.
\begin{document}

\maketitle

\begin{abstract}
Chain-of-thought (CoT) prompting enables reasoning in language models but requires explicit verbalization of intermediate steps. Looped transformers offer an alternative by iteratively refining representations within hidden states. This parameter efficiency comes at a cost, as looped models lack the storage capacity of deeper models which use unique weights per layer. In this work, we investigate transformer models that feature both adaptive per-layer looping, where each transformer block learns to iterate its hidden state via a learned halting mechanism, and gated memory banks, that provide additional learned storage. We find that looping primarily benefits mathematical reasoning, while memory banks help recover performance on commonsense tasks compared to parameter and FLOP matched models. Combining both mechanisms yields a model that outperforms an iso-FLOP baseline---with three times the number of layers---on math benchmarks. Analysis of model internals reveals layer specialization: early layers learn to loop minimally and access memory sparingly, while later layers do both more heavily.

\end{abstract}

\section{Introduction}
Large language models can reason explicitly via chain-of-thought (CoT) prompting \citep{wei2022chain}, which uses step-by-step verbalization to produce reasoning traces, improving performance in downstream tasks. While this is effective, each reasoning step requires generating tokens \citep{nye2021show}, which has motivated interest in \textit{implicit} reasoning, where models perform multi-step computation within their hidden representations without producing intermediate text \citep{saunshi2025reasoning, bae2025mixture}.

One way of implementing implicit reasoning is by stacking transformer layers, iteratively applying the same block which refines the representations through repeated computation. This makes efficient use of parameters---a model that loops $N$ times achieves a larger effective depth without using $N$ times the parameters \citep{graves2016adaptive, dehghani2018universal, banino2021pondernet, goyal2023think}. Recent work has shown that looped transformers can match much deeper non-looped models on reasoning tasks \citep{saunshi2025reasoning, zhu2025ouro, raposo2024mixture}. 

However, a looped model has fundamentally less capacity than a deeper model with $N$ times the number of layers. While loops may improve reasoning, the model has fewer unique parameters in which to encode knowledge. Recent analysis suggests this trade-off is fundamental: looped models achieve their parameter efficiency not through increased knowledge storage but through \emph{knowledge manipulation}---they are able to do multi-hop reasoning while showing similar per-parameter memorization capacity to standard transformers \citep{zhu2025ouro}.

Here, we investigate whether learned memory banks can restore the missing capacity. Specifically, we make the following contributions:
(1) we propose an adapted looped Transformer that combines per-layer adaptive looping with gated access to local and global memory, and
(2) we conduct a systematic study examining the effects of adaptive looping and the inclusion of memory banks on downstream model performance. We find that looping primarily benefits mathematical reasoning, while memory banks help recover commonsense performance compared to parameter- and FLOP-matched models. Analysis of model internals reveals layer specialization: early layers learn to loop minimally and access memory sparingly, while later layers do both more heavily. This specialization means the model learns to choose between thinking harder and knowing more and where to do each.

% Our contributions are as follows: 
% \begin{itemize}
%     \item We propose an adaptive loop model where each layer can be repeated $N$ times and introduce learnable loop scales, which enable the model to gradually learn when and how much to intervene at each iteration. 
%     \item We use local (layer-specific) and global (shared) memory connected via learned gates, and show that looping benefits mathematical reasoning while memory aids knowledge intensive commonsense tasks.
%     \item We analyse model internals and find layer specialization: early layers loop minimally and access memory sparingly, whereas later layers loop more frequently and rely more on memory. We also present preliminary evidence that this loop utilization undergoes a phase transition once basic language competence is achieved.
% \end{itemize}

% ===========================================================================
% FIGURE 1: Architecture
% ===========================================================================
\definecolor{civBlue}{HTML}{00204D}
\definecolor{civBlueMid}{HTML}{414D6B}
\definecolor{civYellow}{HTML}{FFEA5E}
\definecolor{civGold}{HTML}{C7A317}
\begin{figure}[t]
\centering
\resizebox{!}{8cm}{
\begin{tikzpicture}[
    block/.style={
        draw=civBlue,
        fill=civBlue!10,
        minimum width=2cm,
        minimum height=0.8cm,
        rounded corners=3pt,
        font=\small\bfseries,
        text=civBlue
    },
    memblock/.style={
        draw=civGold,
        fill=civYellow!30,
        minimum width=1.4cm,
        minimum height=0.6cm,
        rounded corners=2pt,
        font=\scriptsize,
        text=civBlue!80!black
    },
    gatenode/.style={
        draw=civGold,
        fill=civYellow!50,
        circle,
        minimum size=0.5cm,
        inner sep=0pt,
        font=\large\bfseries,
        text=civBlue
    },
    probnode/.style={
        draw=civBlueMid,
        circle,
        fill=civBlue!5,
        minimum size=0.5cm,
        font=\tiny,
        inner sep=1pt,
        text=civBlue
    },
    hiddennode/.style={font=\small, text=black},
    title/.style={font=\bfseries\small, text=civBlue},
    subtitle/.style={font=\itshape\scriptsize, text=gray},
    arrow/.style={-{Stealth[length=2mm]}, thick, color=civBlue},
    looparrow/.style={-{Stealth[length=2mm]}, thick, color=civBlueMid},
    memarrow/.style={-{Stealth[length=1.5mm]}, color=civGold!90!black, thick},
    probline/.style={-{Stealth[length=1.5mm]}, dashed, color=civBlueMid},
    probconnect/.style={dashed, color=civBlueMid, thick},
]

% SIMPLE MODEL (Left)
\begin{scope}[local bounding box=simple]
    \node[title] at (0, 6.0) {Base Model};
    \node[subtitle] at (0, 5.5) {(Transformer)};
    \node[hiddennode] (h0_s) at (0, 4.5) {$\mathbf{h}^{(0)}$};
    \node[block] (block1_s) at (0, 3.5) {Block 1};
    \node[hiddennode] (h1_s) at (0, 2.5) {$\mathbf{h}^{(1)}$};
    \node[block] (block2_s) at (0, 1.5) {Block 2};
    \node[hiddennode] (h2_s) at (0, 0.5) {$\mathbf{h}^{(2)}$};
    \node[text=civBlueMid] at (0, 0.1) {$\vdots$};
    \node[block] (blockL_s) at (0, -0.7) {Block $L$};
    \node[hiddennode] (hL_s) at (0, -1.7) {$\mathbf{h}^{(L)}$};
    \draw[arrow] (h0_s) -- (block1_s);
    \draw[arrow] (block1_s) -- (h1_s);
    \draw[arrow] (h1_s) -- (block2_s);
    \draw[arrow] (block2_s) -- (h2_s);
    \draw[arrow] (blockL_s) -- (hL_s);
\end{scope}

% LOOP MODEL (Middle)
\begin{scope}[xshift=5.0cm, local bounding box=loop]
    \node[title] at (0, 6.0) {Loop Model};
    \node[subtitle] at (0, 5.5) {(Adaptive Computation)};
    \node[hiddennode] (h0_l) at (0, 4.5) {$\mathbf{h}^{(\ell-1)}$};
    \node[block, minimum height=1.2cm, minimum width=2.4cm] (block_l) at (0, 1.8) {Block $\ell$};
    \node[font=\tiny, text=civGold!70!black, inner sep=1pt] at ($(block_l.south) + (0, 0.25)$) {$\times\zeta(s_n)$};
    \draw[looparrow, rounded corners=6pt]
        ($(block_l.south east) + (-0.3, 0)$) -- ++(0, -0.4) -- ++(0.7, 0) -- ++(0, 0.7)
        node[midway, right, font=\scriptsize, text=civBlueMid, xshift=1pt] {$\times N$}
        -- ($(block_l.east) + (0, -0.2)$);
    \node[draw=civGold, fill=civYellow!20, minimum width=0.8cm, minimum height=0.8cm, font=\scriptsize] (sum_l) at (0, -0.5) {$\sum$};
    \node[font=\tiny, text=gray, right=0.1cm of sum_l] {weighted};
    \node[hiddennode] (hout_l) at (0, -1.7) {$\mathbf{h}^{(\ell)}$};
    \node[probnode] (p1_l) at (-2.0, 2.5) {$p_1$};
    \node[probnode] (p2_l) at (-2.0, 1.6) {$p_2$};
    \node[text=civBlueMid] at (-2.0, 1.1) {$\vdots$};
    \node[probnode] (pN_l) at (-2.0, 0.4) {$p_N$};
    \begin{scope}[on background layer]
        \draw[probline] ($(block_l.north west) + (0.1, 0)$) -- (p1_l.east);
        \draw[probline] ($(block_l.west)$) -- (p2_l.east);
        \draw[probline] ($(block_l.south west) + (0.1, 0)$) -- (pN_l.east);
        \draw[probconnect] (p1_l.south) -- (p2_l.north);
        \draw[probline] (pN_l.south) |- (sum_l.west);
    \end{scope}
    \draw[arrow, civBlue!50] ($(block_l.south) + (-0.6, 0)$) -- ($(sum_l.north) + (-0.2, 0)$);
    \draw[arrow, civBlue!50] (block_l.south) -- (sum_l.north);
    \draw[arrow, civBlue!50] ($(block_l.south) + (0.6, 0)$) -- ($(sum_l.north) + (0.2, 0)$);
    \draw[arrow] (h0_l) -- (block_l);
    \draw[arrow] (sum_l) -- (hout_l);
    \node[font=\tiny, text=gray, align=center] at (-1.2, -0.8) {halt probs};
\end{scope}

% MEMORY MODEL (Right)
\begin{scope}[xshift=10.5cm, local bounding box=memory]
    \node[title] at (0, 6.0) {Memory Model};
    \node[subtitle] at (0, 5.5) {(Adaptive + Memory)};
    \node[memblock, fill=civYellow] (global_mem) at (2.4, 3.8) {Global Mem};
    \node[font=\tiny, text=gray] at (2.4, 3.35) {(shared)};
    \node[memblock] (local_mem) at (2.4, 2.8) {Local Mem};
    \node[font=\tiny, text=gray] at (2.4, 2.35) {(layer $\ell$)};
    \node[hiddennode] (h0_m) at (0, 4.5) {$\mathbf{h}^{(\ell-1)}$};
    \node[gatenode] (gate) at (0, 3.3) {$+$};
    \node[block, minimum height=1.2cm, minimum width=2.4cm] (block_m) at (0, 1.8) {Block $\ell$};
    \node[font=\tiny, text=civGold!70!black, inner sep=1pt] at ($(block_m.south) + (0, 0.25)$) {$\times\zeta(s_n)$};
    \draw[arrow] (h0_m) -- (gate);
    \draw[arrow] (gate) -- (block_m);
    \draw[memarrow] (global_mem.west) -- (gate.east);
    \draw[memarrow] (local_mem.west) -- (gate.east);
    \node[font=\tiny, inner sep=0.5pt, text=civGold!50!black] at (0.95, 3.8) {$\sigma_G$};
    \node[font=\tiny, inner sep=0.5pt, text=civGold!50!black] at (0.95, 2.8) {$\sigma_L$};
    \draw[looparrow, rounded corners=6pt]
        ($(block_m.south east) + (-0.3, 0)$) -- ++(0, -0.4) -- ++(0.7, 0) -- ++(0, 0.7)
        node[midway, right, font=\scriptsize, text=civBlueMid, xshift=1pt] {$\times N$}
        -- ($(block_m.east) + (0, -0.2)$);
    \node[draw=civGold, fill=civYellow!20, minimum width=0.8cm, minimum height=0.8cm, font=\scriptsize] (sum_m) at (0, -0.5) {$\sum$};
    \node[hiddennode] (hout_m) at (0, -1.7) {$\mathbf{h}^{(\ell)}$};
    \node[probnode] (p1_m) at (-2.0, 2.5) {$p_1$};
    \node[probnode] (p2_m) at (-2.0, 1.6) {$p_2$};
    \node[text=civBlueMid] at (-2.0, 1.1) {$\vdots$};
    \node[probnode] (pN_m) at (-2.0, 0.4) {$p_N$};
    \begin{scope}[on background layer]
        \draw[probline] ($(block_m.north west) + (0.1, 0)$) -- (p1_m.east);
        \draw[probline] ($(block_m.west)$) -- (p2_m.east);
        \draw[probline] ($(block_m.south west) + (0.1, 0)$) -- (pN_m.east);
        \draw[probconnect] (p1_m.south) -- (p2_m.north);
        \draw[probline] (pN_m.south) |- (sum_m.west);
    \end{scope}
    \draw[arrow, civBlue!50] ($(block_m.south) + (-0.6, 0)$) -- ($(sum_m.north) + (-0.2, 0)$);
    \draw[arrow, civBlue!50] (block_m.south) -- (sum_m.north);
    \draw[arrow, civBlue!50] ($(block_m.south) + (0.6, 0)$) -- ($(sum_m.north) + (0.2, 0)$);
    \draw[arrow] (sum_m) -- (hout_m);
    \node[font=\tiny, text=gray, align=center] at (-1.2, -0.8) {halt probs};
\end{scope}

% Separating lines
\draw[civBlueMid!30, dashed, thick] (2.4, 6.2) -- (2.4, -2.5);
\draw[civBlueMid!30, dashed, thick] (7.7, 6.2) -- (7.7, -2.5);
\end{tikzpicture}}
\caption{\textbf{Architecture overview.} \emph{Left:} A standard transformer passes hidden states through $L$ unique blocks. \emph{Center:} Our loop model allows each block to iterate up to $N$ times, with a learned halting mechanism that produces a weighted combination of intermediate states. Per-step scales $\zeta(s_n)$ are initialized near zero for training stability. \emph{Right:} The combined model additionally retrieves from local (per-layer) and global (shared) memory banks, gated by learned input-dependent scalars.}
\label{fig:architecture}
\end{figure}
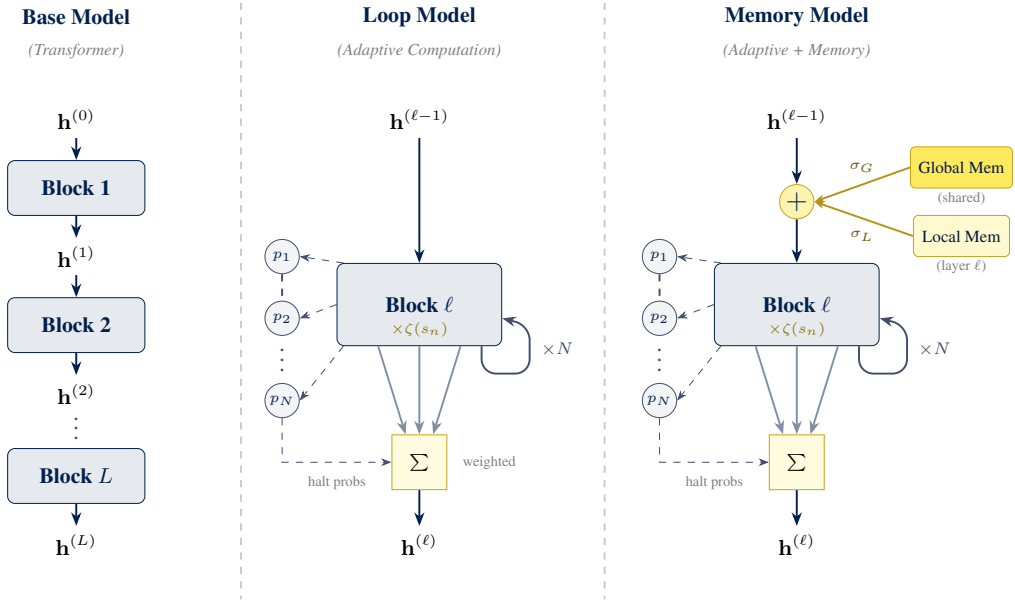

\section{Methods}
We augment a standard decoder-only transformer \citep{vaswani2017attention} with two mechanisms: adaptive looping for repeating computation and memory banks for retrieving learned knowledge. Figure~\ref{fig:architecture} illustrates the architecture and Appendix \ref{app:implementation} provides additional details. 

\subsection{Adaptive Looping}
A standard transformer block applies multi-head self-attention followed by a feed-forward network using residual connections and layer normalization:
\begin{align}
    \mathbf{h}' &= \mathbf{h} + \text{Attn}(\text{LN}(\mathbf{h})) \\
    \mathbf{h}'' &= \mathbf{h}' + \text{FFN}(\text{LN}(\mathbf{h}'))
\end{align}
where $\mathbf{h} \in \mathbb{R}^{B \times T \times D}$ is the hidden state with batch size $B$, sequence length $T$, and embedding dimension $D$. We allow each transformer block to be applied multiple times with a learned halting mechanism, inspired by PonderNet~\citep{banino2021pondernet}. At each iteration $t \in \{1, \ldots, N_{\max}\}$, a halting router predicts the probability of stopping:
\begin{equation}
    p_t = \sigma\left(\mathbf{W}_h \left[\mathbf{h}^{(t)}; t/N_{\max}\right] + b_h\right)
\end{equation}
where $[\cdot; \cdot]$ denotes concatenation and $t/N_{\max}$ provides a normalized step embedding. The final output is computed as a weighted combination over all iterations:
\begin{equation}
    \mathbf{h}_{\text{out}} = \sum_{t=1}^{N_{\max}} p_{\text{halt}}^{(t)} \cdot \mathbf{h}^{(t)}
\end{equation}
where $p_{\text{halt}}^{(t)} = p_t \prod_{i=1}^{t-1}(1 - p_i)$ is the probability of halting at exactly step $t$.

\paragraph{Learnable Loop Scales.} To stabilize model training, we introduce per-step learnable scale parameters. Each iteration applies:
\begin{equation}
    \mathbf{h}^{(t)} = \mathbf{h}^{(t-1)} + \text{softplus}(\alpha_t) \cdot f_\theta(\text{LN}(\mathbf{h}^{(t-1)}))
\end{equation}
where $f_\theta$ denotes the transformer block and $\alpha_t$ is initialized to $-7.0$, which ensures the loop begins as an approximate identity mapping, and the model gradually learns when and how much to intervene.

\subsection{Memory Banks}
\label{sec:memory}
We introduce two types of learned memory, a local and a global one. For the \textit{Local (Per-Layer) Memory} each layer $\ell$ maintains its own memory bank $(\mathbf{K}_\ell, \mathbf{V}_\ell) \in \mathbb{R}^{M_L \times D}$ with $M_L$ slots. This enables layer-specific storage of intermediate computations or specialized knowledge appropriate to that depth. The \textit{Global (Shared) Memory} uses a single memory bank $(\mathbf{K}_G, \mathbf{V}_G) \in \mathbb{R}^{M_G \times D}$ that is shared across all layers, allowing storage of information that might be beneficial for all layers.

Memory retrieval uses scaled dot-product attention with QK-normalization \citep{dehghani2023scaling}:
\begin{align}
    \mathbf{m}_{\text{local}} &= \text{softmax}\!\left(\frac{\text{LN}_q(\mathbf{h}) \cdot \text{LN}_k(\mathbf{K}_\ell)^\top}{\sqrt{D}}\right) \mathbf{V}_\ell \\
    \mathbf{m}_{\text{global}} &= \text{softmax}\!\left(\frac{\text{LN}_q(\mathbf{h}) \cdot \text{LN}_k(\mathbf{K}_G)^\top}{\sqrt{D}}\right) \mathbf{V}_G
\end{align}
Unlike the KV-cache in standard attention, which stores activation history during inference, our memory banks are \textit{static learnable parameters} that are optimized via backpropagation during training but fixed during inference. Our memory implementation draws inspiration from memory-augmented architectures \citep{lample2019large, sukhbaatar2019augmenting, wu2022memorizing} and neural Turing machines~\citep{graves2014neural}.

\paragraph{Gated Memory Integration}
\label{sec:gating}

A critical design choice is how to integrate retrieved memory into the residual stream. Naive addition would force the model to always use memory, potentially harming performance on tasks where loops alone suffice. We therefore employ input-dependent gating:
\begin{align}
    \mathbf{g} &= \sigma(\mathbf{W}_g \mathbf{h} + b_g) \\
    \mathbf{h}_{\text{enriched}} &= \mathbf{h} + \mathbf{g} \odot \mathbf{W}_m \mathbf{m}
\end{align}
where $\odot$ denotes element-wise multiplication. Separate gates control local and global memory contributions:
\begin{equation}
    \mathbf{h}_{\text{memory}} = \mathbf{h} + \mathbf{g}_L \odot \mathbf{W}_L \mathbf{m}_{\text{local}} + \mathbf{g}_G \odot \mathbf{W}_G \mathbf{m}_{\text{global}}
\end{equation}
We study the effect of gate bias initialization $b_g$, comparing $b_g \in \{-3, 0, 3\}$ corresponding to initial gate activations of approximately $\sigma(-3) \approx 0.05$ (nearly closed), $\sigma(0) = 0.5$ (balanced) and $\sigma(3) \approx 0.95$ (nearly open).

\section{Results}

\subsection{Experimental Setup}
\paragraph{Model.} Our base architecture is a decoder-only transformer with $L = 12$ layers and a total of ${\sim}$200M parameters (see Appendix \ref{app:implementation} for full details). For looped models, we use the same 12-layer architecture and allow each layer to iterate up to $N_{\max} \in \{3, 5, 7\}$ times. For memory-augmented models, we add $M_L = $ 1024 local memory slots per layer and $M_G = $ 512 global memory slots, which in total adds approximately 10M parameters. We adapt our iso-param and iso-FLOP models to compensate for the additional parameters from the halting router and per-step scales. We pretrain all models on deduplicated FineWeb-Edu \citep{penedo2024fineweb} for 14B tokens and use a peak learning rate of 0.003. 

\paragraph{Baselines.} We compare against two types of baselines, first a \textbf{Iso-Parameter} model, where the FFN width is increased so that the total parameter count matches the target model. This controls for the possibility that any improvements come simply from having more parameters. And second a \textbf{Iso-FLOP (IsoFLOP)} model, which uses $3{\times}$ the layers (36 layers), matching the forward-pass cost of a model with $N_{\max} = 3$ loops. Table~\ref{tab:ablations} summarizes all configurations. We evaluate on common-sense and math tasks using the OLMES framework ~\citep{gu2025olmes} (see Appendix \ref{app:evaluation} for details).

\begin{table}[t]
\centering
\caption{\textbf{Summary of results}, averaged across benchmarks within each group. CS = commonsense; BPB = bits per byte (lower is better, see Appendix \ref{app:evaluation} for details). Best result per column within each model group is \textbf{bolded}. Full per-benchmark breakdowns are in Appendix \ref{app:fullresults}.}
\label{tab:main_summary}
\setlength{\tabcolsep}{5pt}
\begin{tabular}{@{}llccc@{}}
\toprule
\textbf{Model} & \textbf{Type} & \textbf{CS Acc $\uparrow$} & \textbf{CS BPB $\downarrow$} & \textbf{Math BPB $\downarrow$} \\
\midrule
\multicolumn{5}{l}{\textit{Without Memory}} \\
Base        & IsoPar baseline  & 0.477 & 0.859 & 2.163 \\
Loop-3      & Ours ($N_{\max}{=}3$)   & 0.501 & 0.813 & 1.687 \\
Loop-5      & Ours ($N_{\max}{=}5$)   & 0.503 & 0.823 & 1.737 \\
Loop-7      & Ours ($N_{\max}{=}7$)   & 0.498 & 0.832 & \textbf{1.659} \\
IsoFLOP     & 36-layer         & \textbf{0.523} & \textbf{0.780} & 1.801 \\
\midrule
\multicolumn{5}{l}{\textit{With Memory (all use Loop-3)}} \\
IsoPar-M    & Wider FFN        & 0.459 & 0.823 & 2.108 \\
Mem ($g_0{=}{-}3$)  & Ours (closed init) & 0.472 & 0.810 & 1.619 \\
Mem ($g_0{=}0$)  & Ours (balanced init) & 0.481 & 0.810 & 1.662 \\
Mem ($g_0{=}3$)  & Ours (open init)   & 0.511 & 0.794 & \textbf{1.616} \\
IsoFLOP-M   & 36-layer wider   & \textbf{0.535} & \textbf{0.749} & 1.761 \\
\bottomrule
\end{tabular}
\end{table}

\subsection{Adaptive Loops and Memory}

\paragraph{Looping Improves Mathematical Reasoning}

We first compare averaged benchmark results for looped models without memory (see Table~\ref{tab:main_summary}, full per-benchmark results are given in Appendix \ref{app:fullresults}). Introducing adaptive looping with $N_{\max} = 3$ yields improvements in math BPB (1.687 vs.\ 2.163 for the base model, a 22\% reduction) alongside moderate gains in commonsense accuracy (0.501 vs.\ 0.477) and commonsense BPB (0.813 vs.\ 0.859). Improvements in math are consistent across subcategories with the largest gains on Precalculus ($-31\%$) and Intermediate Algebra ($-26\%$).

When we further increase the number of loops, the performance increase is modest relative to the initial improvement from the base model (Loop-7 improves by 1.7\% over Loop-3). Interestingly, commonsense performance shows a slight downward trend with more loops. These results suggest that additional iterations aid algorithmic computation (math) but do not help tasks that depend on stored knowledge (commonsense). Intriguingly the improvement on math benchmarks remains when we compare against the IsoFLOP model (1.687 vs.\ 1.801, a 6.4\% advantage) despite only having one-third the number of layers. This suggest that looping is a more parameter-efficient way to improve on math benchmarks than simply adding layers, in line with \citet{saunshi2025reasoning}.

\paragraph{Local and Global Memory complements Loops}

We augment the Loop-3 model with local and global memory banks (see Section \ref{sec:memory}) and compare three gate initializations. All memory models share the same architecture and parameter count, only the initial gate bias differs. 

All three memory variants outperform their iso-parameter baseline (IsoPar-M) on both tasks, confirming that the gains are not simply due to having more parameters. Compared to our Loop-3 model without memory we further improve on math benchmarks by 4.2\% and on commonsense accuracy by 2\%, indicating that the memory provides complementary value beyond what looping alone achieves. The comparison to the iso-FLOP baseline shows a similar pattern to above: IsoFLOP-M is better on commonsense but the memory augmented model is better on math benchmarks. Taken together, we observe that memory is able to close some of the commonsense gap that loops alone cannot bridge.

\subsection{Training Dynamics of looped memory Models}

\begin{figure}[t]
    \centering
    \includegraphics[width=\linewidth]{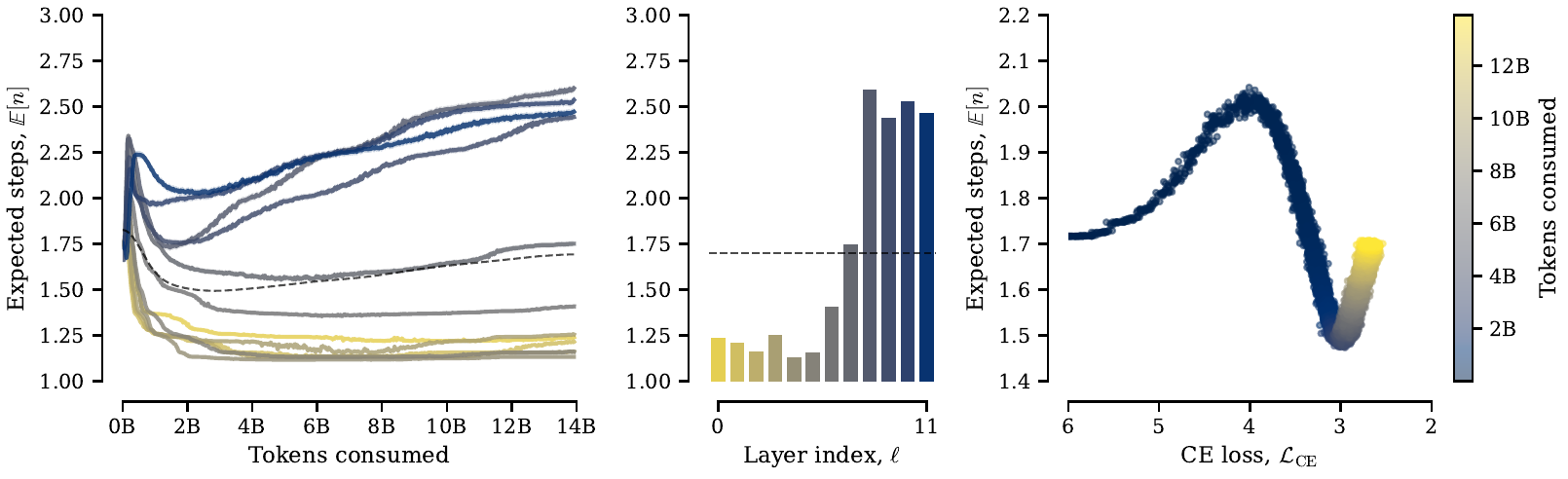}
    \caption{\textbf{Expected number of loop iterations per layer over training.} \emph{Left:} Each curve represents one layer. Early layers (lighter colors) consistently use fewer iterations than later layers (darker colors). \emph{Middle:} Expected steps at the end of training. \emph{Right:} All models show a characteristic transition which occurs at approximately the same cross-entropy value across configurations (see Figure~\ref{fig:ce_transition} in the Appendix for all configurations).}
    \label{fig:layer_computation}
\end{figure}

We further investigated the training dynamics during the training of our models. Since we set $\lambda = 0$ (no ponder penalty), the patterns that emerge are driving entirely by the language modeling objective, i.e. next-token prediction. Figure~\ref{fig:layer_computation} shows the expected number of iterations $\mathbb{E}[n_\ell]$ for each layer $\ell$ over the course of training (see Appendix \ref{app:implementation} for details on $\mathbb{E}[n_\ell]$).

We first observe that not all layers start to loop more over the course of training. We see later layers consistently use more iterations than earlier layers. This seems consistent with studies showing that early transformer layers encode local syntactic patterns while later layers handle more complex semantic and reasoning operations \citep{tenney2019bert, rogers2020primer}. This means the simpler computations performed by early layers do not benefit from iterations while the more complex operations in deeper layers do. 

We also observe that the expected number of loops does not increase monotonically from the start of training (Right side of Figure \ref{fig:layer_computation}). The onset of the increase in the number of loops occurs at approximately the same validation cross-entropy value across all loop configurations, around $3.27 \pm 0.59$ (see Figure \ref{fig:ce_transition} for comparison across models). This suggests the model only begins using additional iterations once it has acquired sufficient language competence to benefit from iterative refinement. 

\section{Discussion}
Our preliminary results point to a functional dissociation between iterative computation and capacity in transformer models. Adaptive looping improves mathematical reasoning but does little for commonsense tasks where additional world knowledge needs to be encoded in the parameters. This aligns with previous work suggesting that transformer feed-forward layers act as key-value memories that store factual associations \citep{geva2021transformer, meng2022locating}, while attention layers route and manipulate information. While looping seems to improve the routing of information, it cannot compensate for insufficient storage capacities. Put differently, the core tradeoff is between knowledge manipulation, which looping enhances as it repeatedly refines the representations, and knowledge capacity, which requires additional unique parameters.

Memory banks are one way of addressing this capacity bottleneck, and when combined with looping show promises in decreasing the gap on commonsense benchmarks. Notably, these dynamics emerge without any ponder penalty. The model is under no explicit pressure to minimize or maximize its loops, therefore the layer-wise specialization we see and the phase transition in the utilization of loops are all consequences of optimizing the language modeling loss alone. 

There are several limitations and open questions which constrain some of the conclusions we can draw. First, our experiments are at a relatively small scale (${\sim}$200M parameters, 12 layers, 14B tokens). Whether our conclusions hold at multi-billion parameter scale, where base models already have substantial capacity, is an open question. Secondly, our math evaluation uses BPB instead of accuracy, which limits our ability to make strong claims about reasoning capabilities. Additionally, while we compare against iso-parameter and iso-FLOP baselines, we do not yet provide a full characterization of the efficiency tradeoff between adding loops or memory slots versus simply increasing depth or width under a continuous compute budget. These limitations will be addressed in follow-up work.

\section{Acknowledgments}
We want to thank Max Lübbering, Timm Heine Ruland, David Fitzek and Richard Rutmann for helpful discussions and technical expertise regarding the Modalities framework \citep{lubbering2026modalities}. This work was funded by the Federal Ministry of Research, Technology \& Space Germany
(BMFTR) and the state of North Rhine-Westphalia
as part of the Lamarr Institute for Machine Learn-
ing and Artificial Intelligence (LAMARR22B).

\newpage
\bibliography{iclr2026_conference}
\bibliographystyle{iclr2026_conference}

\newpage
\appendix
\section{Appendix}

\subsection{Implementation Details}
\label{app:implementation}
Our base models utilize a standard 12-layer transformer architecture with an embedding dimension of $D=768$, $H=12$ attention heads, and an FFN hidden dimension of 3072. With a vocabulary size of 50,304, the total parameter count is approximately 200M. For adaptive looping models, we vary the maximum loop depth $N_{\max} \in \{3, 5, 7\}$ and initialize the loop scale parameter to $\alpha_t = -7.0$. Memory-augmented variants are configured with $M_L = 1024$ local slots and $M_G = 512$ global slots; we ablate gate bias initializations over $b_g \in \{-3.0, 0.0, 3.0\}$.

All models are trained on approximately 13.9B tokens (${\sim}38,620$ steps) using the AdamW optimizer with a batch size of ${\sim}360\text{K}$ tokens. We employ a cosine learning rate schedule with a peak learning rate of $3.0 \times 10^{-3}$. 

For the model loss we combine the next-token prediction loss with an optional ponder penalty:
\begin{equation}
    \mathcal{L} = \mathcal{L}_{\text{CE}} + \lambda \cdot \tilde{n}
\end{equation}
where $\mathcal{L}_{\text{CE}}$ is the categorical cross-entropy and $\tilde{n}$ is the normalized expected number of loop iterations, averaged across all layers:
\begin{equation}
    \tilde{n} = \frac{\bar{n} - 1}{N_{\max} - 1}, \qquad \bar{n} = \frac{1}{L}\sum_{\ell=1}^{L} \mathbb{E}[n_\ell]
\end{equation}
Here $\mathbb{E}[n_\ell]$ denotes the expected step count at layer $\ell$ and $N_{\max}$ is the maximum allowed iterations. This normalization maps the ponder cost to $[0, 1]$, making $\lambda$ interpretable independently of $N_{\max}$.

We set $\lambda = 0$ for the majority of our experiments, meaning the model receives no explicit incentive to minimize loop iterations. Any loop utilization patterns that emerge are driven entirely by the language modeling loss.

\begin{figure}[th]
    \centering
    \includegraphics[width=\linewidth]{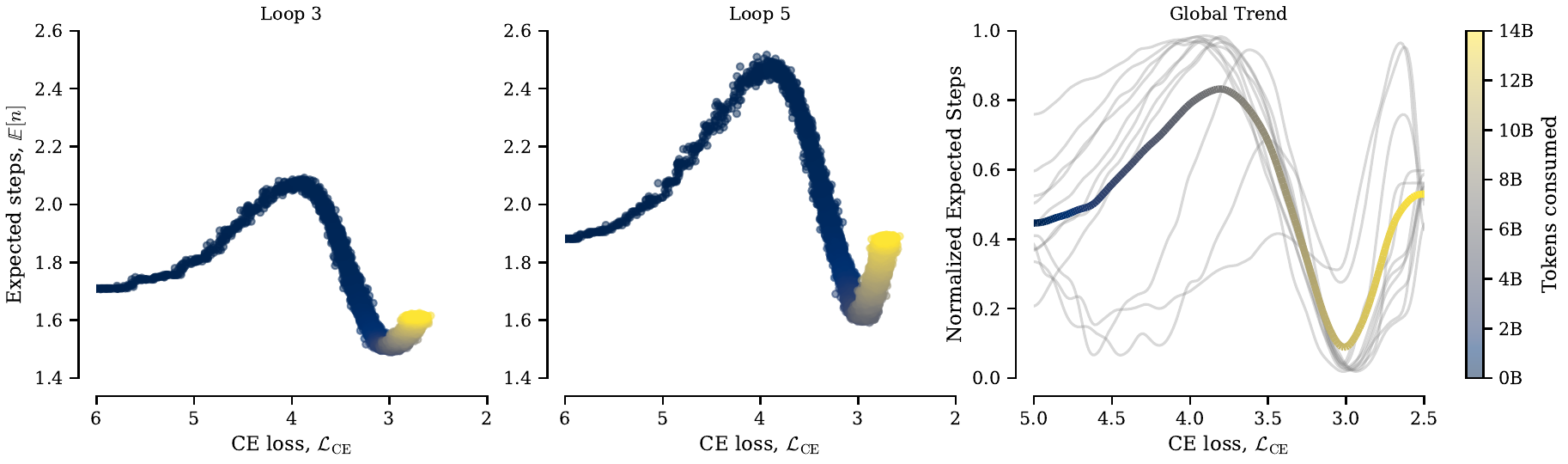}
    \caption{\textbf{Expected loop iterations vs.\ validation cross-entropy for all configurations.} Each point represents one evaluation during training; curves are colored by model configuration. Across all looped models, the expected number of iterations begins to increase rapidly once the cross-entropy drops below approximately $3.27 \pm 0.59$. This phase transition is consistent across Loop-3, Loop-5, and Loop-7 configurations, suggesting it depends on the model's language competence rather than the maximum number of allowed iterations.}
    \label{fig:ce_transition}
\end{figure}

\begin{table}[t]
\centering
\caption{\textbf{Model configurations.} All models use the same base transformer architecture. Loop parameters include per-step scales and halting router weights. Memory parameters include local/global key-value banks and gating networks. Iso-parameter baselines add extra FFN capacity to match the corresponding model's parameter count. Iso-FLOP baselines use 36 layers to approximate the forward-pass cost of 3-loop models.}
\label{tab:ablations}
\setlength{\tabcolsep}{4pt}
\begin{tabular}{@{}lcccl@{}}
\toprule
\textbf{Configuration} & \textbf{Loops} & \textbf{Memory} & \textbf{Parameters} & \textbf{Description} \\
\midrule
IsoPar & \texttimes & \texttimes & 200M & 12 Layers \\
Loop-$N$ & \checkmark & \texttimes & 200M & Adaptive looping, $N_{\max} \in \{3,5,7\}$ \\
IsoFLOP & \texttimes & \texttimes & 332M & 36 Layers \\
\midrule
IsoPar-M & \texttimes & \texttimes & 210M & Wider FFN \\
Mem ($g_0$) & \checkmark & \checkmark & 210M & Loop-3 + Memory \\
IsoFLOP-M & \texttimes & \texttimes & 480M & 36 Layers + Wider FFN \\
\bottomrule
\end{tabular}
\end{table}

\subsection{Evaluation Protocol}
\label{app:evaluation}
We evaluate on two groups of downstream tasks using the OLMes framework~\citep{gu2025olmes}: commonsense benchmarks (ARC-Challenge, ARC-Easy, HellaSwag, LAMBADA, PIQA, QASPER, SocialIQA, Winogrande) and math benchmarks (Algebra, Counting \& Probability, Geometry, Intermediate Algebra, Number Theory, Prealgebra, Precalculus). For commonsense tasks, we report both accuracy and bits-per-byte (BPB). For math tasks, we report BPB only. BPB is computed as the negative log-likelihood of the gold answer divided by the number of UTF-8 bytes in the answer string. Formally, given a negative log-likelihood loss $\ell$, Olmes computes $\text{BPB} = \ell / \ln(2) \cdot (L_T / L_B)$, where $L_T$ is the length in tokens and $L_B$ is the length in UTF-8 bytes \citep{gao2020pile}. Lower BPB indicates better modeling of the target domain. We use BPB as it provides a continuous signal that reveals performance differences throughout pre-training, in contrast to GSM8k, which can remain at or near zero throughout training.

\subsection{Additional Analysis}
\paragraph{Layer-wise specialization of Memory Gates}

\begin{figure}[th]
    \centering
    \includegraphics[width=\linewidth]{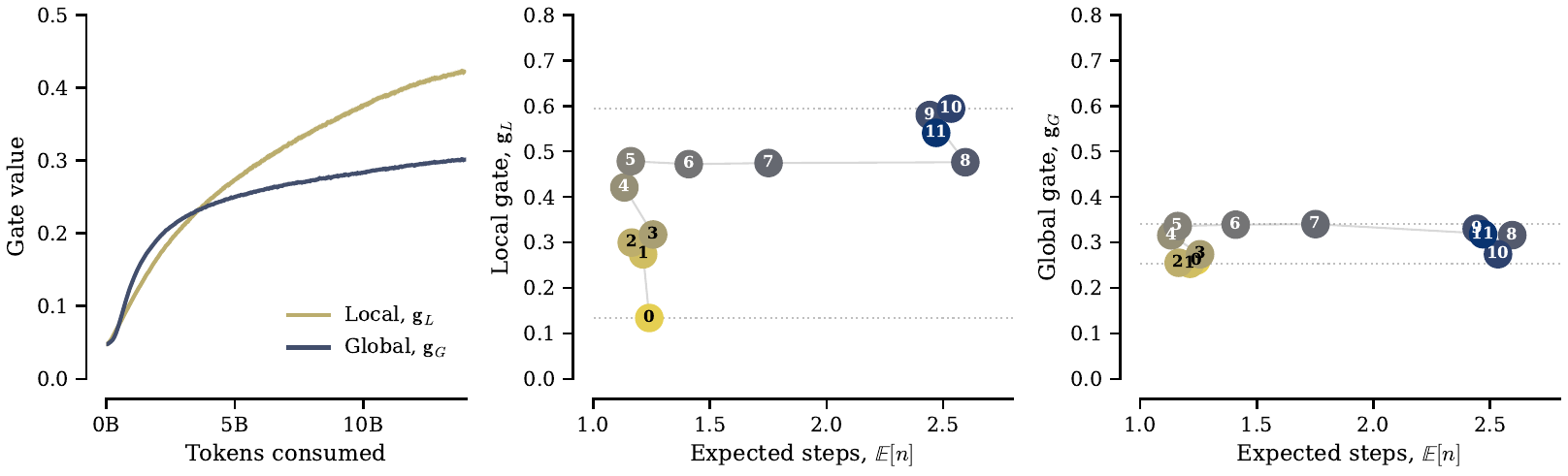}
    \caption{\textbf{Memory gate activations across layers and training.} \emph{Left:} Local memory gate values show high variance across layers while later layers tend to have higher gate activations, and the spread increases over training. \emph{Right:} Global memory gate values increase during training but converge to a more uniform profile across layers, with activations rising up to approximately layer 5 and then plateauing.}
    \label{fig:memory_gates}
\end{figure}

In Figure \ref{fig:memory_gates} we show the dynamics of the local and global memory gate during model training. We observe that local memory gates are used more strongly at the end of training and that the variance across layers is higher than for global gates (local: $0.42 \pm 0.13$, global: $0.30 \pm 0.03$). Since each layers local memory stores distinct key-value pairs, this variance likely reflects differences in the type and amount of information needed at each depth. 

Global memory gates, by contrast, converge to a more uniform activation profile, indicating that the global memory acts as a shared knowledge base, which seems useful at all depths but not requiring layer-specific adaptation. Lastly, we observe that layers that loop more tend to have higher memory gate value, suggesting that the model does not treat loops and memory as substitutes but rather as complements, i.e. layers that need more computation also need more external information. 

\newpage
\subsection{Full Results}
\label{app:fullresults}

% --- FINAL CHECKPOINT ---
\begin{table}[htbp]
\centering
\setlength{\tabcolsep}{2.5pt}
\caption{\textbf{Full results at the final checkpoint.} Best result within each group is bolded. Base = standard transformer; L$N$ = Loop-$N$; L3\textsubscript{IF} = iso-FLOP for Loop-3; M-B\textsubscript{IP} = memory iso-parameter baseline; M\textsubscript{$g_0$} = memory model with gate init $g_0$; M-B\textsubscript{IF} = memory iso-FLOP baseline.}
\label{tab:main_results}
\begin{tabular}{lccccc!{\color{lightgray}\vrule}ccccc}
\toprule
 & \multicolumn{5}{c}{\textbf{Without Memory}} & \multicolumn{5}{c}{\textbf{With Memory}} \\
Bench & Base & L3 & L5 & L7 & L3\textsubscript{IF} & M-B\textsubscript{IP} & M\textsubscript{-3} & M\textsubscript{0} & M\textsubscript{3} & M-B\textsubscript{IF} \\
\midrule
\multicolumn{11}{l}{\textit{Commonsense Accuracy} $\uparrow$} \\
ARC-C & 0.367 & 0.375 & \textbf{0.492} & 0.398 & 0.430 & 0.398 & 0.414 & 0.359 & 0.398 & \textbf{0.438} \\
ARC-E & 0.609 & 0.672 & 0.609 & 0.586 & \textbf{0.688} & 0.602 & 0.648 & 0.672 & 0.641 & \textbf{0.680} \\
HellaSwag & 0.445 & 0.469 & 0.445 & 0.438 & \textbf{0.508} & 0.453 & 0.461 & 0.453 & 0.461 & \textbf{0.508} \\
Lambada & 0.211 & 0.266 & \textbf{0.289} & 0.281 & 0.211 & 0.227 & 0.250 & \textbf{0.281} & 0.242 & 0.273 \\
PIQA & 0.625 & 0.602 & 0.664 & \textbf{0.680} & 0.672 & 0.672 & 0.656 & 0.680 & 0.664 & \textbf{0.688} \\
Qasper & 0.625 & \textbf{0.703} & 0.641 & 0.688 & \textbf{0.703} & 0.422 & 0.531 & 0.484 & 0.680 & \textbf{0.703} \\
SocialIQA & 0.398 & \textbf{0.430} & 0.398 & 0.391 & 0.422 & \textbf{0.445} & 0.398 & 0.430 & 0.406 & 0.438 \\
Winogrande & 0.531 & 0.492 & 0.484 & 0.523 & \textbf{0.555} & 0.453 & 0.414 & 0.484 & \textbf{0.594} & 0.555 \\
\textit{AVG} & 0.477 & 0.501 & 0.503 & 0.498 & \textbf{0.523} & 0.459 & 0.472 & 0.481 & 0.511 & \textbf{0.535} \\
\midrule
\multicolumn{11}{l}{\textit{Commonsense BPB} $\downarrow$} \\
ARC-C & 0.913 & 0.840 & 0.854 & 0.860 & \textbf{0.784} & 0.833 & 0.846 & 0.851 & 0.813 & \textbf{0.754} \\
ARC-E & 0.846 & 0.758 & 0.772 & 0.762 & \textbf{0.706} & 0.789 & 0.740 & 0.733 & 0.721 & \textbf{0.685} \\
HellaSwag & 0.921 & 0.898 & 0.897 & 0.900 & \textbf{0.866} & 0.901 & 0.898 & 0.895 & 0.895 & \textbf{0.849} \\
Lambada & 1.002 & 0.935 & 0.952 & 0.979 & \textbf{0.917} & 0.964 & 0.939 & 0.924 & 0.917 & \textbf{0.847} \\
PIQA & 1.163 & 1.157 & 1.149 & 1.141 & \textbf{1.097} & 1.133 & 1.131 & 1.120 & 1.126 & \textbf{1.064} \\
Qasper & 0.305 & \textbf{0.289} & 0.314 & 0.350 & 0.310 & 0.320 & 0.305 & 0.336 & \textbf{0.294} & 0.295 \\
\textit{AVG} & 0.859 & 0.813 & 0.823 & 0.832 & \textbf{0.780} & 0.823 & 0.810 & 0.810 & 0.794 & \textbf{0.749} \\
\midrule
\multicolumn{11}{l}{\textit{Math BPB} $\downarrow$} \\
Algebra & 2.267 & 1.792 & 1.860 & \textbf{1.766} & 1.895 & 2.244 & 1.718 & 1.773 & \textbf{1.717} & 1.867 \\
Count\&Prob & 1.960 & 1.565 & 1.634 & \textbf{1.530} & 1.641 & 1.946 & 1.491 & 1.524 & \textbf{1.488} & 1.626 \\
Geometry & 1.987 & 1.638 & 1.679 & \textbf{1.618} & 1.717 & 1.930 & 1.577 & 1.613 & \textbf{1.566} & 1.651 \\
IntAlgebra & 2.540 & 1.892 & 1.914 & \textbf{1.839} & 2.067 & 2.481 & \textbf{1.799} & 1.855 & 1.815 & 2.005 \\
NumTheory & 1.855 & 1.560 & 1.588 & \textbf{1.538} & 1.641 & 1.837 & 1.508 & 1.532 & \textbf{1.498} & 1.584 \\
PreAlgebra & 1.755 & 1.437 & 1.482 & \textbf{1.423} & 1.483 & 1.728 & 1.389 & 1.418 & \textbf{1.372} & 1.449 \\
PreCalc & 2.778 & 1.924 & 2.004 & \textbf{1.901} & 2.165 & 2.587 & \textbf{1.847} & 1.917 & 1.854 & 2.147 \\
\textit{AVG} & 2.163 & 1.687 & 1.737 & \textbf{1.659} & 1.801 & 2.108 & 1.619 & 1.662 & \textbf{1.616} & 1.761 \\
\bottomrule
\end{tabular}
\end{table}

% --- EARLY CHECKPOINT ---
\begin{table}[htbp]
\centering
\setlength{\tabcolsep}{2.5pt}
\caption{\textbf{Results at the early checkpoint (step 5000).}}
\label{tab:main_early}
\begin{tabular}{lcccccccccc}
\toprule
Bench & Base & L3 & L5 & L7 & L3\textsubscript{IF} & M-B\textsubscript{IP} & M\textsubscript{-3} & M\textsubscript{0} & M\textsubscript{3} & M-B\textsubscript{IF} \\
\midrule
\multicolumn{11}{l}{\textit{Commonsense Accuracy} $\uparrow$} \\
ARC-C & 0.320 & 0.281 & 0.336 & 0.336 & 0.359 & 0.359 & 0.312 & 0.305 & \textbf{0.367} & 0.344 \\
ARC-E & 0.547 & 0.469 & 0.477 & 0.523 & \textbf{0.602} & 0.531 & 0.508 & 0.508 & 0.555 & 0.586 \\
HellaSwag & \textbf{0.430} & 0.391 & 0.398 & 0.422 & 0.414 & 0.398 & 0.383 & 0.391 & 0.359 & 0.414 \\
Lambada & 0.164 & 0.156 & 0.188 & 0.164 & 0.148 & 0.133 & \textbf{0.219} & 0.203 & 0.164 & 0.172 \\
PIQA & 0.555 & 0.609 & 0.633 & 0.617 & 0.539 & 0.594 & \textbf{0.641} & 0.609 & 0.609 & 0.578 \\
Qasper & 0.664 & \textbf{0.695} & 0.680 & 0.672 & \textbf{0.695} & 0.672 & 0.633 & 0.672 & 0.508 & 0.688 \\
SocialIQA & \textbf{0.461} & 0.430 & 0.398 & 0.414 & 0.398 & 0.422 & 0.406 & 0.422 & 0.414 & 0.445 \\
Winogrande & 0.484 & 0.539 & 0.453 & 0.492 & \textbf{0.586} & 0.531 & 0.531 & 0.531 & 0.477 & 0.531 \\
\textit{AVG} & 0.453 & 0.446 & 0.445 & 0.455 & 0.468 & 0.455 & 0.454 & 0.455 & 0.432 & \textbf{0.470} \\
\midrule
\multicolumn{11}{l}{\textit{Commonsense BPB} $\downarrow$} \\
ARC-C & 1.133 & 1.050 & 1.044 & 1.028 & 1.004 & 1.058 & 1.076 & 1.059 & 1.029 & \textbf{0.980} \\
ARC-E & 1.075 & 0.992 & 0.992 & 0.947 & 0.929 & 1.017 & 1.044 & 0.955 & 0.965 & \textbf{0.891} \\
HellaSwag & 1.022 & 1.017 & 1.004 & 1.005 & 0.986 & 1.014 & 1.007 & 0.996 & 1.000 & \textbf{0.974} \\
Lambada & 1.303 & 1.388 & 1.368 & 1.323 & 1.283 & 1.383 & 1.305 & 1.247 & \textbf{1.232} & 1.270 \\
PIQA & 1.315 & 1.269 & 1.278 & 1.274 & 1.240 & 1.282 & 1.237 & 1.256 & 1.252 & \textbf{1.236} \\
Qasper & 0.357 & \textbf{0.309} & 0.407 & 0.358 & 0.385 & 0.424 & 0.361 & 0.384 & 0.389 & 0.494 \\
\textit{AVG} & 1.034 & 1.004 & 1.015 & 0.989 & \textbf{0.971} & 1.030 & 1.005 & 0.983 & 0.978 & 0.974 \\
\midrule
\multicolumn{11}{l}{\textit{Math BPB} $\downarrow$} \\
Algebra & 2.461 & 2.179 & 2.398 & 2.342 & 2.284 & 2.411 & \textbf{2.038} & 2.067 & 2.093 & 2.295 \\
Count\&Prob & 2.084 & 1.857 & 1.963 & 1.971 & 1.936 & 2.040 & \textbf{1.751} & 1.766 & 1.765 & 1.920 \\
Geometry & 2.186 & 1.970 & 2.068 & 2.090 & 2.032 & 2.135 & \textbf{1.855} & 1.890 & 1.926 & 2.022 \\
IntAlgebra & 2.655 & 2.308 & 2.530 & 2.560 & 2.456 & 2.584 & \textbf{2.168} & 2.174 & 2.231 & 2.457 \\
NumTheory & 2.028 & 1.862 & 1.980 & 1.953 & 1.903 & 2.013 & 1.782 & \textbf{1.770} & 1.782 & 1.890 \\
PreAlgebra & 1.937 & 1.747 & 1.864 & 1.838 & 1.804 & 1.913 & \textbf{1.653} & 1.668 & 1.671 & 1.783 \\
PreCalc & 2.789 & 2.314 & 2.658 & 2.632 & 2.575 & 2.703 & \textbf{2.160} & 2.189 & 2.258 & 2.641 \\
\textit{AVG} & 2.306 & 2.034 & 2.209 & 2.198 & 2.142 & 2.257 & \textbf{1.915} & 1.932 & 1.961 & 2.144 \\
\bottomrule
\end{tabular}
\end{table}

% --- MID CHECKPOINT ---
\begin{table}[htbp]
\centering
\setlength{\tabcolsep}{2.5pt}
\caption{\textbf{Results at the mid checkpoint (step 20000).}}
\label{tab:main_mid}
\begin{tabular}{lcccccccccc}
\toprule
Bench & Base & L3 & L5 & L7 & L3\textsubscript{IF} & M-B\textsubscript{IP} & M\textsubscript{-3} & M\textsubscript{0} & M\textsubscript{3} & M-B\textsubscript{IF} \\
\midrule
\multicolumn{11}{l}{\textit{Commonsense Accuracy} $\uparrow$} \\
ARC-C & 0.398 & 0.414 & \textbf{0.453} & 0.375 & 0.438 & 0.312 & 0.367 & 0.352 & 0.359 & \textbf{0.453} \\
ARC-E & 0.547 & 0.617 & 0.523 & 0.562 & 0.625 & 0.594 & 0.625 & 0.578 & \textbf{0.648} & 0.602 \\
HellaSwag & 0.406 & 0.375 & 0.406 & 0.430 & \textbf{0.461} & 0.391 & 0.422 & 0.438 & 0.430 & 0.453 \\
Lambada & 0.234 & \textbf{0.328} & 0.297 & 0.273 & 0.289 & 0.250 & 0.297 & 0.273 & 0.289 & 0.289 \\
PIQA & 0.594 & 0.625 & 0.609 & 0.656 & 0.664 & 0.633 & 0.641 & 0.641 & 0.656 & \textbf{0.672} \\
Qasper & 0.578 & \textbf{0.688} & 0.375 & \textbf{0.688} & 0.617 & 0.312 & 0.469 & 0.359 & 0.500 & 0.492 \\
SocialIQA & 0.391 & 0.414 & \textbf{0.430} & 0.398 & 0.406 & 0.391 & 0.406 & 0.422 & 0.406 & 0.375 \\
Winogrande & 0.523 & 0.484 & 0.469 & 0.508 & 0.531 & 0.539 & 0.516 & 0.484 & 0.508 & \textbf{0.562} \\
\textit{AVG} & 0.459 & 0.493 & 0.445 & 0.486 & \textbf{0.504} & 0.428 & 0.468 & 0.443 & 0.475 & 0.487 \\
\midrule
\multicolumn{11}{l}{\textit{Commonsense BPB} $\downarrow$} \\
ARC-C & 0.987 & 0.917 & 0.929 & 0.937 & 0.871 & 0.944 & 0.937 & 0.925 & 0.894 & \textbf{0.848} \\
ARC-E & 0.927 & 0.847 & 0.879 & 0.861 & 0.803 & 0.873 & 0.851 & 0.831 & 0.789 & \textbf{0.768} \\
HellaSwag & 0.959 & 0.950 & 0.946 & 0.938 & 0.919 & 0.948 & 0.940 & 0.935 & 0.939 & \textbf{0.906} \\
Lambada & 1.050 & 0.951 & 0.967 & 1.011 & 0.951 & 1.000 & 0.981 & 0.941 & 0.931 & \textbf{0.912} \\
PIQA & 1.217 & 1.216 & 1.195 & 1.204 & 1.152 & 1.203 & 1.186 & 1.187 & 1.173 & \textbf{1.139} \\
Qasper & 0.315 & \textbf{0.300} & 0.347 & 0.324 & 0.332 & 0.453 & 0.355 & 0.352 & 0.320 & 0.333 \\
\textit{AVG} & 0.909 & 0.863 & 0.877 & 0.879 & 0.838 & 0.904 & 0.875 & 0.862 & 0.841 & \textbf{0.818} \\
\midrule
\multicolumn{11}{l}{\textit{Math BPB} $\downarrow$} \\
Algebra & 2.296 & 1.950 & 2.075 & 1.892 & 2.151 & 2.154 & 1.900 & \textbf{1.870} & 1.919 & 2.007 \\
Count\&Prob & 1.983 & 1.677 & 1.808 & 1.622 & 1.822 & 1.874 & 1.626 & \textbf{1.612} & 1.640 & 1.737 \\
Geometry & 2.014 & 1.775 & 1.857 & 1.712 & 1.880 & 1.914 & \textbf{1.701} & 1.712 & 1.719 & 1.778 \\
IntAlgebra & 2.471 & 2.014 & 2.213 & 1.971 & 2.350 & 2.288 & 1.961 & \textbf{1.946} & 2.003 & 2.128 \\
NumTheory & 1.906 & 1.701 & 1.755 & 1.653 & 1.789 & 1.821 & 1.638 & \textbf{1.615} & 1.661 & 1.683 \\
PreAlgebra & 1.820 & 1.567 & 1.643 & \textbf{1.514} & 1.662 & 1.727 & 1.521 & 1.520 & 1.526 & 1.579 \\
PreCalc & 2.597 & 2.045 & 2.343 & 1.990 & 2.492 & 2.375 & \textbf{1.979} & 1.984 & 2.043 & 2.252 \\
\textit{AVG} & 2.155 & 1.819 & 1.956 & 1.765 & 2.021 & 2.022 & 1.761 & \textbf{1.751} & 1.787 & 1.880 \\
\bottomrule
\end{tabular}
\end{table}

\end{document}